%% file: main-1742-Artetxe.tex
\documentclass[11pt,a4paper]{article}
\usepackage{times,latexsym}
\usepackage{url}
\usepackage[T1]{fontenc}

\usepackage[acceptedWithA]{tacl2018v2}

\usepackage[utf8]{inputenc}
\usepackage{graphicx}
\usepackage{booktabs} %
\usepackage{multirow}
\usepackage{amsmath}
\DeclareMathOperator{\score}{score}
\DeclareMathOperator{\nn}{NN}
\DeclareMathOperator{\margin}{margin}

\newcommand{\NbLangTrain}{93 }
\newcommand{\NbLangTrainGen}{29 }
\newcommand{\NbLangTest}{112 }
\newcommand{\NbScript}{28 }
\newcommand{\bpeN}{50k }

\input{tables.tex}

\title{Massively Multilingual Sentence Embeddings for Zero-Shot Cross-Lingual Transfer and Beyond}

\author{Mikel Artetxe \\
  University of the Basque Country (UPV/EHU)\thanks{This work was performed during an internship at Facebook AI Research.} \\
  {\sf mikel.artetxe@ehu.eus} \\
  \And
  Holger Schwenk \\
  Facebook AI Research \\
  {\sf schwenk@fb.com} \\}

\date{}

\begin{document}
\maketitle
\begin{abstract}
We introduce an architecture to learn joint multilingual sentence representations for \NbLangTrain languages, belonging to more than 30 different families and written in \NbScript different scripts. Our system uses a single \mbox{BiLSTM} encoder with a shared BPE vocabulary for all languages, which is coupled with an auxiliary decoder and trained on publicly available parallel corpora. This enables us to learn a classifier on top of the resulting embeddings using English annotated data only, and transfer it to any of the \NbLangTrain languages without any modification.
Our experiments in cross-lingual natural language inference (XNLI dataset), cross-lingual document classification (MLDoc dataset) and parallel corpus mining (BUCC dataset) show the effectiveness of our approach. We also introduce a new test set of aligned sentences in \NbLangTest languages, and show that our sentence embeddings obtain strong results in multilingual similarity search even for low-resource languages. Our implementation, the pre-trained encoder and the multilingual test set are available at \url{https://github.com/facebookresearch/LASER}.
\end{abstract}

\section{Introduction}
\label{sec:introduction}

While the recent advent of deep learning has led to impressive progress in Natural Language Processing (NLP), these techniques are known to be particularly data hungry, limiting their applicability in many practical scenarios. An increasingly popular approach to alleviate this issue is to first learn general language representations on unlabeled data, which are then integrated in task-specific downstream systems. This approach was first popularized by word embeddings \citep{mikolov2013distributed,pennington2014glove}, but has recently been superseded by sentence-level representations \citep{peters2018deep,devlin2018bert}. Nevertheless, all these works learn a separate model for each language and are thus unable to leverage information across different languages, greatly limiting their potential performance for low-resource languages.

In this work, we are interested in \textbf{universal language agnostic sentence embeddings}, that is, vector representations of sentences that are general with respect to two dimensions: the input language and the NLP task. The motivations for such representations are multiple: the hope that languages with limited resources benefit from joint training over many languages, the desire to perform zero-shot transfer of an NLP model from one language (typically English) to another, and the possibility to handle code-switching. To that end, we train a single encoder to handle multiple languages, so that semantically similar sentences in different languages are close in the  embedding space.

While previous work in multilingual NLP has been limited to either a few languages \citep{schwenk2017learning,yu2018multilingual} or specific applications like typology prediction \citep{malaviya2017learning} or machine translation \citep{neubig2018rapid}, we learn general purpose sentence representations for \NbLangTrain languages (see Table \ref{tab:languages}). Using a single pre-trained BiLSTM encoder for all the \NbLangTrain languages, we obtain very strong results in various scenarios without any fine-tuning, including cross-lingual natural language inference (XNLI dataset), cross-lingual classification (MLDoc dataset), bitext mining (BUCC dataset) and a new multilingual similarity search dataset we introduce covering 112 languages. To the best of our knowledge, this is the first exploration of general purpose massively multilingual sentence representations across a large variety of tasks.

\section{Related work}
\label{sec:related}

Following the success of word embeddings \citep{mikolov2013distributed,pennington2014glove}, there has been an increasing interest in learning continuous vector representations of longer linguistic units like sentences \citep{le2014distributed,kiros2015skipthought}. These sentence embeddings are commonly obtained using a Recurrent Neural Network (RNN) encoder, which is typically trained in an unsupervised way over large collections of unlabelled corpora. For instance, the skip-thought model of \citet{kiros2015skipthought} couple the encoder with an auxiliary decoder, and train the entire system to predict the surrounding sentences over a collection of books. It was later shown that more competitive results could be obtained by training the encoder over labeled Natural Language Inference (NLI) data \citep{conneau2017supervised}. This was later extended to multitask learning, combining different training objectives like that of skip-thought, NLI and machine translation \citep{cer2018universal,subramanian2018learning}.

While the previous methods consider a single language at a time, multilingual representations have attracted a large attention in recent times. Most of this research focuses on cross-lingual word embeddings \citep{ruder2017survey}, which are commonly learned jointly from parallel corpora \citep{gouws2015bilbowa,luong2015bilingual}. An alternative approach that is becoming increasingly popular is to separately train word embeddings for each language, and map them to a shared space based on a bilingual dictionary \citep{mikolov2013exploiting,artetxe2018generalizing} or even in a fully unsupervised manner \cite{conneau2018word,artetxe2018robust}. Cross-lingual word embeddings are often used to build bag-of-word representations of longer linguistic units by taking their respective (IDF-weighted) average \citep{klementiev2012inducing,dufter2018embedding}. While this approach has the advantage of requiring weak or no cross-lingual signal, it has been shown that the resulting sentence embeddings work poorly in practical cross-lingual transfer settings \citep{conneau2018xnli}.

A more competitive approach that we follow here is to use a sequence-to-sequence encoder-decoder architecture \citep{schwenk2017learning,hassan2018achieving}. The full system is trained end-to-end on parallel corpora akin to multilingual neural machine translation \citep{johnson2017google}: the encoder maps the source sequence into a fixed-length vector representation, which is used by the decoder to create the target sequence. This decoder is then discarded, and the encoder is kept to embed sentences in any of the training languages. While some proposals use a separate encoder for each language \citep{schwenk2017learning}, sharing a single encoder for all languages also gives strong results \citep{schwenk2018filtering}.

Nevertheless, most existing work is either limited to few, rather close languages \citep{schwenk2017learning,yu2018multilingual} or, more commonly, consider pairwise joint embeddings with English and one foreign language \citep{espana2017empirical,guo2018effective}. To the best of our knowledge, existing work on learning multilingual representations for a large number of languages is limited to word embeddings \citep{ammar2016massively,dufter2018embedding} or specific applications like typology prediction \citep{malaviya2017learning} or machine translation \citep{neubig2018rapid}, ours being the first paper exploring general purpose massively multilingual sentence representations.

All the previous approaches learn a fixed-length representation for each sentence. A recent research line has obtained very strong results using variable-length representations instead, consisting of contextualized embeddings of the words in the sentence \citep{dai2015semisupervised,peters2018deep,howard2018universal,devlin2018bert}. For that purpose, these methods train either an RNN or self-attentional encoder over unnanotated corpora using some form of language modeling. A classifier can then be learned on top of the resulting encoder, which is commonly further fine-tuned during this supervised training. Concurrent to our work, \citet{lample2019cross} propose a cross-lingual extension of these models, and report strong results in cross-lingual natural language inference, machine translation and language modeling. In contrast, our focus is on scaling to a large number of languages, for which we argue that fixed-length approaches provide a more versatile and compatible representation form.\footnote{For instance, there is not always a one-to-one correspondence among words in different languages (e.g. a single word of a morphologically complex language might correspond to several words of a morphologically simple language), so having a separate vector for each word might not transfer as well across languages.} Also, our approach achieves strong results without task-specific fine-tuning, which makes it interesting for tasks with limited resources.

\begin{figure*}[t]
  \centering
  \includegraphics[width=0.95\textwidth]{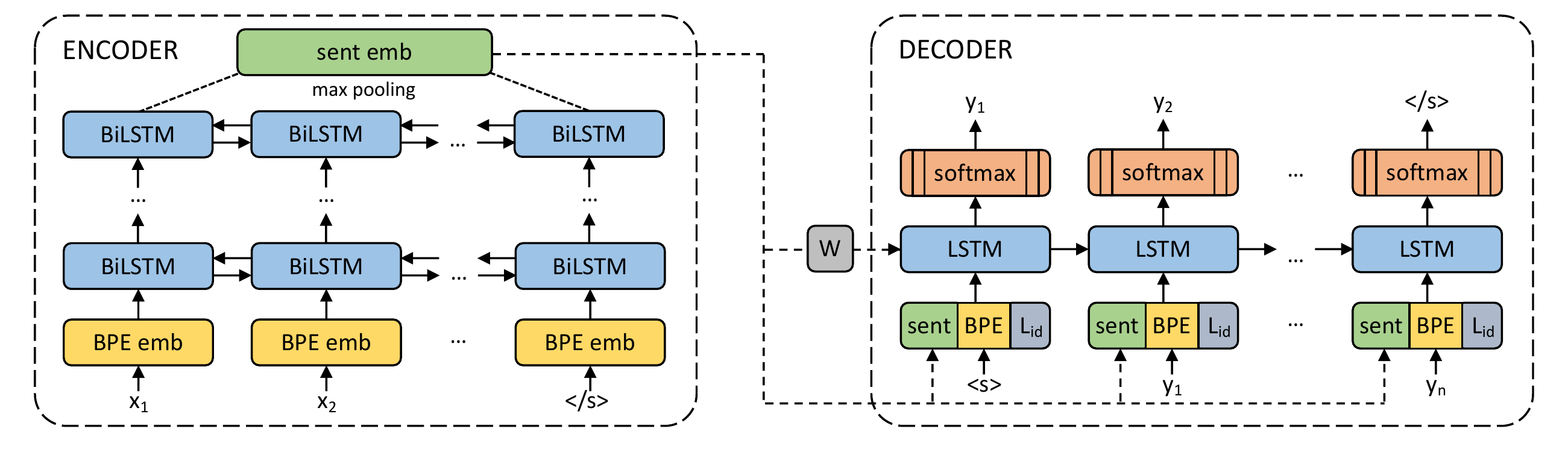}
  \caption{Architecture of our system to learn multilingual sentence embeddings.}
  \label{fig:architecture}
\end{figure*}

\section{Proposed method}
\label{sec:embeddings}

We use a single, language agnostic BiLSTM encoder to build our sentence embeddings, which is coupled with an auxiliary decoder and trained on parallel corpora. From Section \ref{subsec:architecture} to \ref{subsec:data}, we describe its architecture, our training strategy to scale to \NbLangTrain languages, and the training data used for that purpose.

\subsection{Architecture} \label{subsec:architecture}

Figure \ref{fig:architecture} illustrates the architecture of the proposed system, which is based on \citet{schwenk2018filtering}.
As it can be seen, sentence embeddings are obtained by applying a max-pooling operation over the output of a BiLSTM encoder. These sentence embeddings are used to initialize the decoder LSTM through a linear transformation, and are also concatenated to its input embeddings at every time step. Note that there is no other connection between the encoder and the decoder, as we want all relevant information of the input sequence to be captured by the sentence embedding.

We use a single encoder and decoder in our system, which are shared by all languages involved. For that purpose, we build a joint byte-pair encoding (BPE) vocabulary with \bpeN operations, which is learned on the concatenation of all training corpora. This way, the encoder has no explicit signal on what the input language is, encouraging it to learn language independent representations. In contrast, the decoder takes a language ID embedding that specifies the language to generate, which is concatenated to the input and sentence embeddings at every time step.

Scaling up to almost one hundred languages calls for an encoder with sufficient capacity. In this paper, we limit our study to a stacked BiLSTM with 1 to 5 layers, each 512-dimensional. The resulting sentence representations (after concatenating both directions) are 1024 dimensional. The decoder has always one layer of dimension 2048. The input embedding size is set to 320, while the language ID embedding has 32 dimensions.

\begin{table*}[t!]
  \insertTabLanguages
  \caption{List of the \NbLangTrain languages along with their training size, the resulting similarity error rate on Tatoeba, and the number of sentences in it. Dashes denote language pairs excluded for containing less than 100 test sentences.}
  \label{tab:languages}
\end{table*}

\subsection{Training strategy} \label{subsec:strategies}

In preceding work \cite{schwenk2017learning,schwenk2018filtering}, each input sentence was jointly translated into all other languages. However, this approach has two obvious drawbacks when trying to scale to a large number of languages. First, it requires an N-way parallel corpus, which is difficult to obtain for all languages. Second, it has a quadratic cost with respect to the number of languages, making training prohibitively slow as the number of languages is increased. In our preliminary experiments, we observed that similar results can be obtained using only two target languages.\footnote{Note that, if we had a single target language, the only way to train the encoder for that language would be auto-encoding, which we observe to work poorly. Having two target languages avoids this problem.} At the same time, we relax the requirement for N-way parallel corpora by considering separate alignments for each language combination.

Training minimizes the cross-entropy loss on the training corpus, alternating over all combinations of the languages involved. For that purpose, we use Adam with a constant learning rate of 0.001 and dropout set to 0.1, and train for a fixed number of epochs. Our implementation is based on \texttt{fairseq},\footnote{\url{https://github.com/pytorch/fairseq}} and we make use of its multi-GPU support to train on 16 NVIDIA V100 GPUs with a total batch size of 128,000 tokens. Unless otherwise specified, we train our model for 17 epochs, which takes about 5 days. Stopping training earlier decreases the overall performance only slightly.

\begin{table*}[t]
  \insertTabXNLI
  \caption{Test accuracies on the XNLI cross-lingual natural language inference dataset. All results from \citet{conneau2018xnli} correspond to max-pooling, which outperforms the last-state variant in all cases. Results involving MT do not use a multilingual model and are not directly comparable with zero-shot transfer. Overall best results are in bold, the best ones in each group are underlined. \\
  $^*$ Results for BERT \cite{devlin2018bert} are extracted from its GitHub README\footnotemark[9] \\ %
  $^\dagger$ Monolingual BERT model for Thai from \url{https://github.com/ThAIKeras/bert}
  }
  \label{tab:results_xnli}
\end{table*}

\subsection{Training data and pre-processing}
\label{subsec:data}

As described in Section \ref{subsec:strategies}, training requires bitexts aligned with two target languages. We choose English and Spanish for that purpose, as most of the data is aligned with these languages.\footnote{Note that it is not necessary that all input languages are systematically aligned with both target languages. Once we have several languages with both alignments, the joint embedding is well conditioned, and we can add more languages with one alignment only, usually English.} We collect training corpora for \NbLangTrain input languages by combining the Europarl, United Nations, OpenSubtitles2018, Global Voices, Tanzil and Tatoeba corpus, which are all publicly available on the OPUS website\footnote{\url{http://opus.nlpl.eu}} \cite{tiedmann2012parallel}. Appendix~\ref{app:data} provides a more detailed description of this training data, while Table~\ref{tab:languages} summarizes the list of all languages covered and the size of the bitexts. Our training data comprises a total of 223 million parallel sentences. All pre-processing is done with Moses tools:\footnote{\url{http://www.statmt.org/moses}} punctuation normalization, removing non-printing characters and tokenization. As the only exception, Chinese and Japanese were segmented with Jieba\footnote{\url{https://github.com/fxsjy/jieba}} and Mecab,\footnote{\url{https://github.com/taku910/mecab}} respectively. All the languages are kept in their original script with the exception of Greek, which we romanize into the Latin alphabet. It is important to note that the joint encoder itself has no information on the language or writing script of the tokenized input texts. It is even possible to mix multiple languages in one sentence.

\section{Experimental evaluation}
\label{sec:experiments}

In contrast with the well-established evaluation frameworks for English sentence representations \cite{conneau2017supervised,wang2018glue}, there is not yet a commonly accepted standard to evaluate multilingual sentence embeddings. The most notable effort in this regard is arguably the XNLI dataset \cite{conneau2018xnli}, which evaluates the transfer performance of an NLI model trained on English data over 14 additional test languages (Section \ref{subsec:xnli}). So as to obtain a more complete picture, we also evaluate our embeddings in cross-lingual document classification (MLDoc,  Section \ref{subsec:mldoc}), and bitext mining (BUCC, Section \ref{subsec:bucc}). However, all these datasets only cover a subset of our \NbLangTrain languages, so we also introduce a new test set for multilingual similarity search in \NbLangTest languages, including several languages for which we have no training data but whose language family is covered (Section \ref{subsec:tatoeba}).
We remark that we use the same pre-trained BiLSTM encoder for all tasks and languages without any fine-tuning.

\subsection{XNLI: cross-lingual NLI}
\label{subsec:xnli}

\begin{table*}[t]
  \insertTabMLDoc
  \caption{Accuracies on the MLDoc zero-shot cross-lingual document classification task (test set).}
  \label{tab:results_mldoc}
\end{table*}

NLI has become a widely used task to evaluate sentence representations \citep{bowman2015large,williams2018broad}. Given two sentences, a premise and a hypothesis, the task consists in deciding whether there is an \textit{entailment}, \textit{contradiction} or \textit{neutral} relationship between them. XNLI is a recent effort to create a dataset similar to the English MultiNLI for several languages \cite{conneau2018xnli}.
It consists of 2,500 development and 5,000 test instances translated from English into 14 languages by professional translators, making results across different languages directly comparable.

We train a classifier on top of our multilingual encoder using the usual combination of the two sentence embeddings: $(p, h, p \cdot h, |p-h|)$, where $p$ and $h$ are the premise and hypothesis. For that purpose, we use a feed-forward neural network with two hidden layers of size 512 and 384, trained with Adam. All hyperparameters were optimized on the English XNLI development corpus only, and then, the same classifier was applied to all languages of the XNLI test set. As such, we did not use any training or development data in any of the foreign languages. Note, moreover, that the multilingual sentence embeddings are fixed and not fine-tuned on the task or the language.

We report our results in Table~\ref{tab:results_xnli}, along with several baselines from \citet{conneau2018xnli} and the multilingual BERT model \citep{devlin2018bert}.\footnote{\label{foot:bert}Note that the multilingual variant of BERT is not discussed in its paper \citep{devlin2018bert}. Instead, the reported results were extracted from the README of the official GitHub project at \url{https://github.com/google-research/bert/blob/master/multilingual.md} on July 5, 2019.} Our proposed method obtains the best results in zero-shot cross-lingual transfer for all languages but Spanish. Moreover, our transfer results are strong and homogeneous across all languages: for 11 of them, the zero-short performance is (at most) 5\% lower than the one on English, including distant languages like Arabic, Chinese and Vietnamese, and we also achieve remarkable good results on low-resource languages like Swahili. In contrast, BERT achieves excellent results on English, outperforming our system by 7.5 points, but its transfer performance is much weaker. For instance, the loss in accuracy for both Arabic and Chinese is 2.5 points for our system, compared to 19.3 and 17.6 points for BERT.\footnote{Concurrent to our work, \citet{lample2019cross} report superior results using another variant of BERT, outperforming our method by 4.5 points in average. However, note that these results are not fully comparable because 1) their system uses development data in the foreign languages, whereas our approach is fully zero-shot, 2) their approach requires fine-tuning on the task, 3) our system handles a much larger number of languages, and 4) our transfer performance is substantially better (an average loss of 4 vs 10.6 points with respect to the respective English system).} Finally, we also outperform all baselines of \citet{conneau2018xnli} by a substantial margin, with the additional advantage that we use a single pre-trained encoder, whereas \mbox{X-BiLSTM} learns a separate encoder for each language.

We also provide results involving Machine Translation (MT) from \citet{conneau2018xnli}. This can be done in two ways: 1) translate the test data into English and apply the English NLI classifier, or 2) translate the English training data and train a separate NLI classifier for each language. Note that we are not evaluating multilingual sentence embeddings anymore, but rather the quality of the MT system and a monolingual model. Moreover, the use of MT incurs in an important overhead with either strategy: translating test makes inference substantially more expensive, whereas translating train results in a separate model for each language. As shown in Table~\ref{tab:results_xnli}, our approach outperforms all translation baselines of \citet{conneau2018xnli}. We also outperform MT BERT for Arabic and Thai, and are very close for Urdu. Thanks to its multilingual nature, our system can also handle premises and hypothesis in different languages. As reported in Appendix \ref{app:xnli_cross}, the proposed method obtains very strong results in these settings, even for distant language combinations like French-Chinese.

\subsection{MLDoc: cross-lingual classification}
\label{subsec:mldoc}

\begin{table*}[t!]
  \insertTabBucc
  \caption{F1 scores on the BUCC mining task.}
  \label{tab:results_bucc}
\end{table*}

Cross-lingual document classification is a typical application of multilingual representations. In order to evaluate our sentence embeddings in this task, we use the MLDoc dataset of \citet{schwenk2018corpus}, which is an improved version of the Reuters benchmark \citep{lewis2004rcv1,klementiev2012inducing} with uniform class priors and a wider language coverage. There are 1,000 training and development documents and 4,000 test documents for each language, divided in 4 different genres. Just as with the XNLI evaluation, we consider the zero-shot transfer scenario: we train a classifier on top of our multilingual encoder using the English training data, optimizing hyper-parameters on the English development set, and evaluating the resulting system in the remaining languages. We use a feed-forward neural network with one hidden layer of 10 units.

As shown in Table~\ref{tab:results_mldoc}, our system obtains the best published results for 5 of the 7 transfer languages. We believe that our weaker performance on Japanese can be attributed to the domain and sentence length mismatch between MLDoc and the parallel corpus we use for this language.

\subsection{BUCC: bitext mining} \label{subsec:bucc}

Bitext mining is another natural application for multilingual sentence embeddings. Given two comparable corpora in different languages, the task consists in identifying sentence pairs that are translations of each other. For that purpose, one would commonly score sentence pairs by taking the cosine similarity of their respective embeddings, so parallel sentences can be extracted through nearest neighbor retrieval and filtered by setting a fixed threshold over this score \citep{schwenk2018filtering}. However, it was recently shown that this approach suffers from scale inconsistency issues \citep{guo2018effective}, and \citet{artetxe2018margin} proposed the following alternative score addressing it:
\begin{multline*}
    \score(x, y) = \margin (\cos(x, y), \\
    \sum_{z \in \nn_k(x)}{\frac{\cos(x, z)}{2k}} +  \sum_{z \in \nn_k(y)}{\frac{\cos(y, z)}{2k}})
\end{multline*}
where $x$ and $y$ are the source and target sentences, and $\nn_k(x)$ denotes the $k$ nearest neighbors of $x$ in the other language. The paper explores different margin functions, with \textit{ratio} ($\margin(a, b) = \frac{a}{b}$) yielding the best results. This notion of margin is related to CSLS \citep{conneau2018word}.

We use this method to evaluate our sentence embeddings on the BUCC mining task \citep{zweigenbaum2017overview,zweigenbaum2018overview}, using exact same hyper-parameters as \citet{artetxe2018margin}. The task consists in extracting parallel sentences from a comparable corpus between English and four foreign languages: German, French, Russian and Chinese. The dataset consists of 150K to 1.2M sentences for each language, split into a sample, training and test set, with about 2--3\% of the sentences being parallel. As shown in Table \ref{tab:results_bucc}, our system establishes a new state-of-the-art for all language pairs with the exception of English-Chinese test. We also outperform \citet{artetxe2018margin} themselves, who use two separate models covering 4 languages each. Not only are our results better, but our model also covers many more languages, so it can potentially be used to mine bitext for any combination of the \NbLangTrain languages supported.

\subsection{Tatoeba: similarity search}
\label{subsec:tatoeba}

While XNLI, MLDoc and BUCC are well established benchmarks with comparative results available, they only cover a small subset of our \NbLangTrain languages. So as to better assess the performance of our model in all these languages, we introduce a new test set of similarity search for \NbLangTest languages based on the Tatoeba corpus. The dataset consists of up to 1,000 English-aligned sentence pairs for each language. Appendix~\ref{app:tatoeba} describes how the dataset was constructed in more details. Evaluation is done by finding the nearest neighbor for each sentence in the other language according to cosine similarity and computing the error rate.

We report our results in Table~\ref{tab:languages}. Contrasting these results with those of XNLI, one would assume that similarity error rates below 5\% are indicative of strong downstream performance.\footnote{We consider the average of en$\rightarrow$xx and xx$\rightarrow$en} This is the case for 37 languages, while there are 48 languages with an error rate below 10\% and 55 with less than 20\%. There are only 15 languages with error rates above 50\%. Additional result analysis is given in Appendix \ref{app:tatoeba_analysis}.

We believe that our competitive results for many low-resource languages are indicative of the benefits of joint training, which is also supported by our ablation results in Section \ref{subsec:ablation_languages}. In relation to that, Appendix~\ref{app:unseen} reports similarity search results for \NbLangTrainGen additional languages without any training data, showing that our encoder can also generalize to unseen languages to some extent as long as it was trained on related languages.

\section{Ablation experiments}
\label{sec:ablation}

In this section, we explore different variants of our approach and study the impact on the performance for all our evaluation tasks. We report average results across all languages. For XNLI, we also report the accuracy on English.

\subsection{Encoder depth}

Table~\ref{tab:results:depth} reports the performance on the different tasks for encoders with 1, 3 or 5 layers. We were not able to achieve good convergence with deeper models. It can be seen that all tasks benefit from deeper models, in particular XNLI and Tatoeba, suggesting that a single layer BiLSTM has not enough capacity to encode so many languages.

\begin{table}[t]
  \insertTabDepth
  \caption{Impact of the depth of the BiLSTM encoder.}
  \label{tab:results:depth}
\end{table}

\subsection{Multitask learning}

Multitask learning has been shown to be helpful to learn English sentence embeddings \cite{subramanian2018learning,cer2018universal}. The most important task in this approach is arguably NLI, so we explored adding an additional NLI objective to our system with different weighting schemes. As shown in Table \ref{tab:results:nli}, the NLI objective leads to a better performance on the English NLI test set, but this comes at the cost of a worse cross-lingual transfer performance in XNLI and Tatoeba. The effect in BUCC is negligible.

\begin{table}[t]
  \insertTabNLI
  \caption{Multitask training with an NLI objective and different weightings.}
  \label{tab:results:nli}
\end{table}

\subsection{Number of training languages}
\label{subsec:ablation_languages}

So as to better understand how our architecture scales to a large amount of languages, we train a separate model on a subset of 18 evaluation languages, and compare it to our main model trained on \NbLangTrain languages. We replaced the Tatoeba corpus with the WMT 2014 test set to evaluate the multilingual similarity error rate. This covers English, Czech, French, German and Spanish, so results between both models are directly comparable. As shown in Table~\ref{tab:results:nb_langs}, the full model equals or outperforms the one covering the evaluation languages only for all tasks but MLDoc. This suggests that the joint training also yields to overall better representations.

\begin{table}[t]
  \insertTabNLangs
  \caption{Comparison between training on \NbLangTrain languages and training on the 18 evaluation languages only.}
  \label{tab:results:nb_langs}
\end{table}

\section{Conclusions}
\label{sec:conclusions}

In this paper, we propose an architecture to learn multilingual fixed-length sentence embeddings for \NbLangTrain languages. We use a single language-agnostic \mbox{BiLSTM} encoder for all languages, which is trained on publicly available parallel corpora and applied to different downstream tasks without any fine-tuning. Our experiments on cross-lingual natural language inference (XNLI), cross-lingual document classification (MLDoc), and bitext mining (BUCC) confirm the effectiveness of our approach. We also introduce a new test set of multilingual similarity search in \NbLangTest languages, and show that our approach is competitive even for low-resource languages.
To the best of our knowledge, this is the first successful exploration of general purpose massively multilingual sentence representations.

In the future, we would like to explore alternative encoder architectures like self-attention \citep{vaswani2017attention}. We would also like to explore strategies to exploit monolingual data, such as using pre-trained word embeddings, back-translation \citep{sennrich2016improving,edunov2018understanding}, or other ideas from unsupervised MT \citep{artetxe2018unsupervised,lample2018phrase}. Finally, we would like to replace our language dependant pre-processing with a language agnostic approach like SentencePiece.\footnote{\url{https://github.com/google/sentencepiece}}

Our implementation, the pre-trained encoder and the multilingual test set are freely available at \url{https://github.com/facebookresearch/LASER}.

\newpage
\bibliography{tacl2018}
\bibliographystyle{acl_natbib}
\newpage

\appendix

\section{Training data}
\label{app:data}

Our training data consists of the combination of the following publicly available parallel corpora:
\begin{itemize}
    \item \textbf{Europarl}: 21 European languages. The size varies from 400k to 2M sentences depending on the language pair.
    \item \textbf{United Nations}: We use the first two million sentences in Arabic, Russian and Chinese.
    \item \textbf{OpenSubtitles2018:} A parallel corpus of movie subtitles in 57 languages. The corpus size varies from a few thousand sentences to more than 50 million. We keep at most 2 million entries for each language pair.
    \item \textbf{Global Voices:} News stories from the Global Voices website (38 languages). This is a rather small corpus with less than 100k sentence in most of the languages.
    \item \textbf{Tanzil:} Quran translations in 42 languages, average size of 135k sentences. The style and vocabulary is very different from news texts.
    \item \textbf{Tatoeba:} A community supported collection of English sentences and translations into more than 300 languages. We use this corpus to extract a separate test set of up to 1,000 sentences (see Appendix~\ref{app:tatoeba}). For languages with more than 1,000 entries, we use the remaining ones for training.
\end{itemize}

Using all these corpora would provide parallel data for more languages, but we decided to keep \NbLangTrain languages after discarding several constructed languages with little practical use (Klingon, Kotava, Lojban, Toki Pona and Volapük). In our preliminary experiments, we observed that the domain of the training data played a key role in the performance of our sentence embeddings. Some tasks (BUCC, MLDoc) tend to perform better when the encoder is trained on long and formal sentences, whereas other tasks (XNLI, Tatoeba) benefit from training on shorter and more informal sentences. So as to obtain a good balance, we used at most two million sentences from OpenSubtitles, although more data is available for some languages. The size of the available training data varies largely for the considered languages (see Table~\ref{tab:languages}). This favours high-resource languages when the joint BPE vocabulary is created and the training of the joint encoder. In this work, we did not try to counter this effect by over-sampling low-resource languages.

\section{XNLI results for all language combinations}
\label{app:xnli_cross}

\begin{table*}[t!]
  \insertTabXNLIcross
  \caption{XNLI test accuracies for our approach when the premise and hypothesis are in different languages.}
  \label{tab:xnli_cross}
\end{table*}

\begin{table*}[t!]
  \insertTabTatoebaGen
  \caption{Performance on the Tatoeba test set for languages for which we have no training data.}
  \label{tab:languagesGen}
\end{table*}

Table~\ref{tab:xnli_cross} reports the accuracies of our system on the XNLI test set when the premises and hypothesis are in a different language. The numbers in the diagonal correspond to the main results reported in Table \ref{tab:results_xnli}. Our approach obtains strong results when combining different languages. We do not have evidence that distant languages perform considerably worse. Instead, the combined performance seems mostly bounded by the accuracy of the language that performs worst when used alone. For instance, Greek-Russian achieves very similar results to Bulgarian-Russian, two Slavic languages. Similarly, combing French with Chinese, two totally different languages, is only 1.5 points worse than French-Spanish, two very close languages.

\section{Tatoeba: dataset}
\label{app:tatoeba}

Tatoeba\footnote{\url{https://tatoeba.org/eng/}} is an open collection of English sentences and high-quality translations into more than 300 languages. The number of available translations is updated every Saturday. We downloaded the snapshot on November 19th 2018 and performed the following processing:
    1) removal of sentences containing ``@'' or ``http'', as emails and web addresses are not language specific;
    2) removal of sentences with less than three words, as they usually have little semantic information;
    3) removal of sentences that appear multiple times, either in the source or the target.

After filtering, we created test sets of up to 1,000 aligned sentences with English. This amount is available for 72 languages. Limiting the number of sentences to 500, we increase the coverage to 86 languages, and 112 languages with 100 parallel sentences. It should be stressed that, in general, the English sentences are not the same for different languages, so error rates are not directly comparable across languages.

\section{Tatoeba: result analysis}
\label{app:tatoeba_analysis}

In this section, we provide some analysis on the results given in Table \ref{tab:languages}. We have 48 languages with an error rate below 10\% and 55 with less than 20\%, respectively (English included). The languages with less than 20\% error belong to 20 different families and use 12 different scripts, and include 6 languages for which we have only small amounts of bitexts (less than 400k), namely Esperanto, Galician, Hindi, Interlingua, Malayam and Marathi, which presumably benefit from the joint training with other related languages.

Overall, we observe low similarity error rates on the Indo-Aryan languages, namely Hindi, Bengali, Marathi and Urdu. The performance on Berber languages (``ber'' and ``kab'') is remarkable, although we have less than 100k sentences to train them. This is a typical example of languages which are spoken by several millions of people, but for which the amount of written resources is very limited.  It is quite unlikely that we would be able to train a good sentence embedding with language specific corpora only, showing the benefit of joint training on many languages.

Only 15 languages have similarity error rates above 50\%. Four of them are low-resource languages with their own script and which are alone in their family (Amharic, Armenian, Khmer and Georgian), making it difficult to benefit from joint training. In any case, it is still remarkable that a language like Khmer performs much better than random with only 625 training examples. There are also several Turkic languages (Kazakh, Tatar, Uighur and Uzbek) and Celtic languages (Breton and Cornish) with high error rates. We plan to further investigate its cause and possible solutions in the future.

\section{Tatoeba: results for unseen languages}
\label{app:unseen}

We extend our experiments to \NbLangTrainGen languages without any training data (see Table~\ref{tab:languagesGen}). Many of them are recognized minority languages spoken in specific regions (e.g. Asturian, Faroese, Frisian, Kashubian, North Moluccan Malay, Piemontese, Swabian or Sorbian). All share some similarities, at various degrees, with other major languages that we cover, but also differ by their own grammar or specific vocabulary.  This enables our encoder to perform reasonably well, even if it did not see these languages during training.

\end{document}

%% file: tables.tex
\newcommand{\insertTabLanguages}{
  \begin{center}
  \begin{footnotesize}
  \addtolength{\tabcolsep}{-4pt}
  \begin{tabular}{lccccccccccccccccccccccc}
    \toprule
\bf  & \bf af & \bf am & \bf ar & \bf ay & \bf az & \bf be & \bf ber & \bf bg & \bf bn & \bf br & \bf bs & \bf ca & \bf cbk & \bf cs & \bf da & \bf de \\
\bf train sent. & 67k & 88k & 8.2M & 14k & 254k & 5k & 62k & 4.9M & 913k & 29k & 4.2M & 813k & 1k & 5.5M & 7.9M & 8.7M \\
\bf en$\rightarrow$xx err. & 11.20 & 60.71 & 8.30 & n/a & 44.10 & 31.20 & 29.80 & 4.50 & 10.80 & 83.50 & 3.95 & 4.00 & 24.20 & 3.10 & 3.90 & 0.90 \\
\bf xx$\rightarrow$en err. & 9.90 & 55.36 & 7.80 & n/a & 23.90 & 36.50 & 33.70 & 5.40 & 10.00 & 84.90 & 3.11 & 4.20 & 21.70 & 3.80 & 4.00 & 1.00 \\
\bf test sent. & 1000 & 168 & 1000 & -- & 1000 & 1000 & 1000 & 1000 & 1000 & 1000 & 354 & 1000 & 1000 & 1000 & 1000 & 1000 \\
\midrule
\bf  & \bf dtp & \bf dv & \bf el & \bf en & \bf eo & \bf es & \bf et & \bf eu & \bf fi & \bf fr & \bf ga & \bf gl & \bf ha & \bf he & \bf hi & \bf hr \\
\bf train sent. & 1k & 90k & 6.5M & 2.6M & 397k & 4.8M & 5.3M & 1.2M & 7.9M & 8.8M & 732 & 349k & 127k & 4.1M & 288k & 4.0M \\
\bf en$\rightarrow$xx err. & 92.10 & n/a & 5.30 & n/a & 2.70 & 1.90 & 3.20 & 5.70 & 3.70 & 4.40 & 93.80 & 4.60 & n/a & 8.10 & 5.80 & 2.80 \\
\bf xx$\rightarrow$en err. & 93.50 & n/a & 4.80 & n/a & 2.80 & 2.10 & 3.40 & 5.00 & 3.70 & 4.30 & 95.80 & 4.40 & n/a & 7.60 & 4.80 & 2.70 \\
\bf test sent. & 1000 & -- & 1000 & -- & 1000 & 1000 & 1000 & 1000 & 1000 & 1000 & 1000 & 1000 & -- & 1000 & 1000 & 1000 \\
\midrule
\bf  & \bf hu & \bf hy & \bf ia & \bf id & \bf ie & \bf io & \bf is & \bf it & \bf ja & \bf ka & \bf kab & \bf kk & \bf km & \bf ko & \bf ku & \bf kw \\
\bf train sent. & 5.3M & 6k & 9k & 4.3M & 3k & 3k & 2.0M & 8.3M & 3.2M & 296k & 15k & 4k & 625 & 1.4M & 50k & 2k \\
\bf en$\rightarrow$xx err. & 3.90 & 59.97 & 5.40 & 5.20 & 14.70 & 17.40 & 4.40 & 4.60 & 3.90 & 60.32 & 39.10 & 80.17 & 77.01 & 10.60 & 80.24 & 91.90 \\
\bf xx$\rightarrow$en err. & 4.00 & 67.79 & 4.10 & 5.80 & 12.80 & 15.20 & 4.40 & 4.80 & 5.40 & 67.83 & 44.70 & 82.61 & 81.72 & 11.50 & 85.37 & 93.20 \\
\bf test sent. & 1000 & 742 & 1000 & 1000 & 1000 & 1000 & 1000 & 1000 & 1000 & 746 & 1000 & 575 & 722 & 1000 & 410 & 1000 \\
\midrule
\bf  & \bf kzj & \bf la & \bf lfn & \bf lt & \bf lv & \bf mg & \bf mhr & \bf mk & \bf ml & \bf mr & \bf ms & \bf my & \bf nb & \bf nds & \bf nl & \bf oc \\
\bf train sent. & 560 & 19k & 2k & 3.2M & 2.0M & 355k & 1k & 4.2M & 373k & 31k & 2.9M & 2k & 4.1M & 12k & 8.4M & 3k \\
\bf en$\rightarrow$xx err. & 91.60 & 41.60 & 35.90 & 4.10 & 4.50 & n/a & 87.70 & 5.20 & 3.35 & 9.00 & 3.40 & n/a & 1.30 & 18.60 & 3.10 & 39.20 \\
\bf xx$\rightarrow$en err. & 94.10 & 41.50 & 35.10 & 3.40 & 4.70 & n/a & 91.50 & 5.40 & 2.91 & 8.00 & 3.80 & n/a & 1.10 & 15.60 & 4.30 & 38.40 \\
\bf test sent. & 1000 & 1000 & 1000 & 1000 & 1000 & -- & 1000 & 1000 & 687 & 1000 & 1000 & -- & 1000 & 1000 & 1000 & 1000 \\
\midrule
\bf  & \bf pl & \bf ps & \bf pt & \bf ro & \bf ru & \bf sd & \bf si & \bf sk & \bf sl & \bf so & \bf sq & \bf sr & \bf sv & \bf sw & \bf ta & \bf te \\
\bf train sent. & 5.5M & 4.9M & 8.3M & 4.9M & 9.3M & 91k & 796k & 5.2M & 5.2M & 85k & 3.2M & 4.0M & 7.8M & 173k & 42k & 33k \\
\bf en$\rightarrow$xx err. & 2.00 & 7.20 & 4.70 & 2.50 & 4.90 & n/a & n/a & 3.10 & 4.50 & n/a & 1.80 & 4.30 & 3.60 & 45.64 & 31.60 & 18.38 \\
\bf xx$\rightarrow$en err. & 2.40 & 6.00 & 4.90 & 2.70 & 5.90 & n/a & n/a & 3.70 & 3.77 & n/a & 2.30 & 5.00 & 3.20 & 39.23 & 29.64 & 22.22 \\
\bf test sent. & 1000 & 1000 & 1000 & 1000 & 1000 & -- & -- & 1000 & 823 & -- & 1000 & 1000 & 1000 & 390 & 307 & 234 \\
\midrule
\bf  & \bf tg & \bf th & \bf tl & \bf tr & \bf tt & \bf ug & \bf uk & \bf ur & \bf uz & \bf vi & \bf wuu & \bf yue & \bf zh \\
\bf train sent. & 124k & 4.1M & 36k & 5.7M & 119k & 88k & 1.4M & 746k & 118k & 4.0M & 2k & 4k & 8.3M \\
\bf en$\rightarrow$xx err. & n/a & 4.93 & 47.40 & 2.30 & 72.00 & 59.90 & 5.80 & 20.00 & 82.24 & 3.40 & 25.80 & 37.00 & 4.10 \\
\bf xx$\rightarrow$en err. & n/a & 4.20 & 51.50 & 2.60 & 65.70 & 49.60 & 5.10 & 16.20 & 80.37 & 3.00 & 25.20 & 38.90 & 5.00 \\
\bf test sent. & -- & 548 & 1000 & 1000 & 1000 & 1000 & 1000 & 1000 & 428 & 1000 & 1000 & 1000 & 1000 \\
  \bottomrule
  \end{tabular}
  \end{footnotesize}
  \end{center}
}

\newcommand{\insertTabTatoebaGen}{
  \begin{center}
  \begin{footnotesize}
  \addtolength{\tabcolsep}{-4pt}
  \begin{tabular}{lcccccccccccccccccccccc}
    \toprule
\bf  & \bf ang & \bf arq & \bf arz & \bf ast & \bf awa & \bf ceb & \bf ch & \bf csb & \bf cy & \bf dsb & \bf fo & \bf fy & \bf gd & \bf gsw & \bf hsb \\
\bf en$\rightarrow$xx err. & 58.96 & 58.62 & 31.24 & 12.60 & 63.20 & 81.67 & 64.23 & 54.55 & 89.74 & 48.64 & 28.24 & 46.24 & 95.66 & 52.99 & 42.44 \\
\bf xx$\rightarrow$en err. & 65.67 & 62.46 & 31.03 & 14.96 & 64.50 & 87.00 & 77.37 & 58.89 & 93.04 & 55.32 & 28.63 & 50.29 & 96.98 & 58.12 & 48.65 \\
\bf test sent. & 134 & 911 & 477 & 127 & 231 & 600 & 137 & 253 & 575 & 479 & 262 & 173 & 829 & 117 & 483 \\
\midrule
\bf  & \bf jv & \bf max & \bf mn & \bf nn & \bf nov & \bf orv & \bf pam & \bf pms & \bf swg & \bf tk & \bf tzl & \bf war & \bf xh & \bf yi \\
\bf en$\rightarrow$xx err. & 73.66 & 48.24 & 89.55 & 13.40 & 33.07 & 68.26 & 93.10 & 50.86 & 50.00 & 75.37 & 54.81 & 84.20 & 90.85 & 93.28 \\
\bf xx$\rightarrow$en err. & 80.49 & 50.00 & 94.09 & 10.00 & 35.02 & 75.45 & 95.00 & 49.90 & 58.04 & 83.25 & 55.77 & 88.60 & 92.25 & 95.40 \\
\bf test sent. & 205 & 284 & 440 & 1000 & 257 & 835 & 1000 & 525 & 112 & 203 & 104 & 1000 & 142 & 848 \\
  \bottomrule
  \end{tabular}
  \end{footnotesize}
  \end{center}
}

\newcommand{\insertTabXNLI}{
  \begin{center}
  \begin{footnotesize}
  \addtolength{\tabcolsep}{-3.5pt}
  \begin{tabular}{llclcccccccccccccc}
    \toprule
    & & \multirow{2}{*}{EN} & & \multicolumn{14}{c}{EN $\rightarrow$ XX} \\
    \cmidrule{5-18}
    & & & & fr & es & de & el & bg & ru & tr & ar & vi & th & zh & hi & sw & ur \\
    \midrule
    \multicolumn{16}{l}{\bf Zero-Shot Transfer, one NLI system for all languages:} \\[2pt]
    \multirow{2}{2.0cm}{\citet{conneau2018xnli}} 
    & X-BiLSTM & 73.7 & & 
     67.7 & 68.7 & 67.7 & 68.9 & 67.9 & 65.4 & 64.2 & 64.8 & 66.4 & 64.1 & 65.8 & 64.1 & 55.7 & 58.4 \\
    & X-CBOW & 64.5 & &
     60.3 & 60.7 & 61.0 & 60.5 & 60.4 & 57.8 & 58.7 & 57.5 & 58.8 & 56.9 & 58.8 & 56.3 & 50.4 & 52.2 \\
    BERT uncased$^*$ & Transformer & \underline{81.4} & &
     -- & \underline{74.3} & 70.5 & -- & -- & -- & -- & 62.1 & -- & -- & 63.8 & -- & -- & 58.3 \\
    \midrule
    Proposed method & BiLSTM 
    & 73.9 & &
      \bf 71.9 &     72.9 & \underline{72.6} & {\bf 72.8} & {\bf 74.2} & {\bf 72.1} & {\bf 69.7} &
      {\bf 71.4} & {\bf 72.0} & {\bf 69.2} & \underline{71.4} & {\bf 65.5} & {\bf 62.2} & \underline{61.0} \\[10pt]
      
    \toprule
    \multicolumn{16}{l}{\bf Translate test, one English NLI system:} \\[2pt]
    \citet{conneau2018xnli} & BiLSTM & 73.7 & &
      \underline{70.4} & 70.7 & 68.7 & \underline{69.1} & \underline{70.4} & \underline{67.8} & \underline{66.3} & 66.8 & \underline{66.5} & 64.4 & 68.3 & \underline{64.2} & \underline{61.8} & 59.3 \\
    BERT uncased$^*$ & Transformer & 81.4 & &
      -- &  74.9 & 74.4 & -- & -- & -- & -- & 70.4 & -- & -- & 70.1 & -- & -- & \bf 62.1 \\[10pt]
    \multicolumn{16}{l}{\bf Translate train, separate NLI systems for each language:} \\[2pt]
    \citet{conneau2018xnli} & BiLSTM & 73.7 & &
      68.3 & 68.8 & 66.5 & 66.4 & 67.4 & 66.5 & 64.5 & 65.8 & 66.0 & 62.8 & 67.0 & 62.1 & 58.2 & 56.6 \\
    BERT cased$^*$ & Transformer & \bf 81.9 & &
      -- &  \bf 77.8 & \bf 75.9 & -- & -- & -- & -- & \underline{70.7} & -- & \underline{\it 68.9}$^\dagger$ & \bf 76.6 & -- & -- & 61.6 \\
    \bottomrule
  \end{tabular}
  \end{footnotesize}
  \end{center}
}

\newcommand{\insertTabXNLIcross}{
  \begin{center}
  \begin{footnotesize}
  \addtolength{\tabcolsep}{-3.5pt}
  \begin{tabular}{r*{15}{cc@{\hspace{7pt}}}@{\hspace{10pt}}c}
\toprule
&& \multicolumn{15}{c}{Hypothesis} \\
&&  en  &   ar  &   bg  &   de  &   el  &   es  &   fr  &   hi  &   ru  &   sw  &   th  &   tr  &   ur  &   vi  &   zh  &   avg \\
\midrule
 \multirow{15}{*}{\rotatebox{90}{Premise}}
&en & \bf 73.9 & 70.0 & 72.0 & 72.8 & 71.6 & 72.2 & 72.2 & 65.9 & 71.4 & 61.5 & 67.6 & 69.7 & 61.0 & 70.7 & 70.3 & 69.5 \\
&ar & 70.5 & \bf 71.4 & 71.1 & 70.1 & 69.6 & 70.6 & 70.0 & 64.9 & 69.9 & 60.1 & 67.1 & 68.2 & 60.6 & 69.5 & 70.1 & 68.2 \\
&bg & 72.7 & 71.1 & \bf 74.2 & 72.3 & 71.7 & 72.1 & 72.7 & 65.5 & 71.7 & 60.8 & 69.0 & 69.8 & 61.2 & 70.5 & 70.5 & 69.7 \\
&de & 72.0 & 69.6 & 71.8 & \bf 72.6 & 70.9 & 71.7 & 71.5 & 65.2 & 70.8 & 60.5 & 68.1 & 69.1 & 60.5 & 70.0 & 70.7 & 69.0 \\
&el & 73.0 & 70.1 & 72.0 & 72.4 & \bf 72.8 & 71.5 & 71.9 & 65.2 & 71.7 & 61.0 & 68.1 & 69.5 & 61.0 & 70.2 & 70.4 & 69.4 \\
&es & 73.3 & 70.4 & 72.4 & 72.7 & 71.5 & \bf 72.9 & 72.2 & 65.0 & 71.2 & 61.5 & 68.1 & 69.8 & 60.5 & 70.4 & 70.4 & 69.5 \\
&fr & 73.2 & 70.4 & 72.2 & 72.5 & 71.1 & 72.1 & \bf 71.9 & 65.9 & 71.3 & 61.4 & 68.1 & 70.0 & 60.9 & 70.9 & 70.4 & 69.5 \\
&hi & 66.7 & 66.0 & 66.7 & 67.2 & 65.4 & 66.1 & 65.6 & \bf 65.5 & 66.5 & 58.9 & 63.8 & 65.9 & 59.5 & 65.6 & 66.0 & 65.0 \\
&ru & 71.3 & 70.0 & 72.3 & 71.4 & 70.5 & 71.2 & 71.3 & 64.4 & \bf 72.1 & 60.8 & 67.9 & 68.7 & 60.5 & 69.9 & 70.1 & 68.8 \\
&sw & 65.7 & 64.5 & 65.7 & 65.0 & 65.1 & 65.2 & 64.5 & 61.5 & 64.9 & \bf 62.2 & 63.3 & 64.5 & 58.2 & 65.0 & 65.1 & 64.0 \\
&th & 70.5 & 69.2 & 71.4 & 70.1 & 69.6 & 70.2 & 69.6 & 65.2 & 70.2 & 62.1 & \bf 69.2 & 67.7 & 60.9 & 70.0 & 69.6 & 68.4 \\
&tr & 70.6 & 69.1 & 70.4 & 70.3 & 69.6 & 70.6 & 69.8 & 64.0 & 69.1 & 61.3 & 67.3 & \bf 69.7 & 60.6 & 69.8 & 69.0 & 68.1 \\
&ur & 65.5 & 64.8 & 65.3 & 65.9 & 65.3 & 65.7 & 64.8 & 62.1 & 65.3 & 58.2 & 63.2 & 64.1 & \bf 61.0 & 64.3 & 65.0 & 64.0 \\
&vi & 71.7 & 69.7 & 72.2 & 71.1 & 70.7 & 71.3 & 70.5 & 65.4 & 71.0 & 61.3 & 69.0 & 69.3 & 60.6 & \bf 72.0 & 70.3 & 69.1 \\
&zh & 71.6 & 69.9 & 71.7 & 71.1 & 70.1 & 71.2 & 70.8 & 64.1 & 70.9 & 60.5 & 68.6 & 68.9 & 60.3 & 69.8 & \bf 71.4 & 68.7 \\
&avg & 70.8 & 69.1 & 70.8 & 70.5 & 69.7 & 70.3 & 70.0 & 64.7 & 69.8 & 60.8 & 67.2 & 68.3 & 60.5 & 69.2 & 69.3 & \bf 68.1 \\
\bottomrule
\end{tabular}
  \end{footnotesize}
  \end{center}
}

\newcommand{\insertTabMLDoc}{
  \begin{center}
  \begin{footnotesize}
  \addtolength{\tabcolsep}{-2pt}
  \begin{tabular}{llclccccccc}
    \toprule
    & & \multirow{2}{*}{EN} & & \multicolumn{7}{c}{EN $\rightarrow$ XX} \\
    \cmidrule{5-11}
    & & & & de & es & fr & it & ja & ru & zh \\
    \midrule
    \multirow{3}{1.15cm}{\citet{schwenk2018corpus}}
    & MultiCCA + CNN & \bf 92.20 & & 81.20 & 72.50 & 72.38 & 69.38 & \bf 67.63 & 60.80 & \bf 74.73 \\
    & BiLSTM (Europarl) & 88.40 & & 71.83 & 66.65 & 72.83 & 60.73 & - & - & - \\
    & BiLSTM (UN) & 88.83 & & - & 69.50 & 74.52 & - & - & 61.42 & 71.97 \\
    \midrule
    \multicolumn{2}{l}{Proposed method} & 89.93 & & 
    \bf 84.78 & \bf 77.33 & \bf 77.95 & \bf 69.43 & 60.30 & \bf 67.78 & 71.93 \\
    \bottomrule
  \end{tabular}
  \end{footnotesize}
  \end{center}
}

\newcommand{\insertTabBucc}{
  \begin{center}
  \begin{footnotesize}
  \addtolength{\tabcolsep}{-1pt}
  \begin{tabular}{llllllllll}
    \toprule
    & \multicolumn{4}{c}{TRAIN} & & \multicolumn{4}{c}{TEST} \\
    \cmidrule{2-5} \cmidrule{7-10}
    & \multicolumn{1}{c}{de-en} & \multicolumn{1}{c}{fr-en} & \multicolumn{1}{c}{ru-en} & \multicolumn{1}{c}{zh-en} & & \multicolumn{1}{c}{de-en} & \multicolumn{1}{c}{fr-en} & \multicolumn{1}{c}{ru-en} & \multicolumn{1}{c}{zh-en} \\
    \midrule
    \citet{azpeitia2017weighted} & 83.33 & 78.83 & - & - & & 83.74 & 79.46 & - & - \\
    \citet{gregoire2017bucc} & - & 20.67 & - & - & & - & 20 & - & - \\
    \citet{zhang2017znlp} & - & - & - & 43.48 & & - & - & - & 45.13 \\
    \citet{azpeitia2018extracting} & 84.27 & 80.63 & 80.89 & 76.45 & & 85.52 & 81.47 & 81.30 & 77.45 \\
    \citet{bouamor2018h2} & - & 75.2 & - & - & & - & 76.0 & - & - \\
    \citet{chongman2018neural} & - & - & - & 58.54 & & - & - & - & 56 \\
    \citet{schwenk2018filtering} & 76.1 & 74.9 & 73.3 & 71.6 & & 76.9 & 75.8 & 73.8 & 71.6 \\
    \citet{artetxe2018margin} & 94.84 & 91.85 & 90.92 & 91.04 & & 95.58 & 92.89 & 92.03 & \bf 92.57 \\
    \midrule
    Proposed method
    & \bf 95.43 & \bf 92.40 & \bf 92.29 & \bf 91.20 & & \bf 96.19 & \bf 93.91 & \bf 93.30 & 92.27 \\
    \bottomrule
  \end{tabular}
  \end{footnotesize}
  \end{center}
 }

\newcommand{\insertTabDepth}{
  \begin{center}
  \begin{footnotesize}
  \addtolength{\tabcolsep}{-3pt}
  \begin{tabular}{c*{7}{c}}
    \toprule
    \multirow{2}{*}{Depth}
     & Tatoeba & BUCC & MLDoc & XNLI-en & XNLI-xx \\
     & Err [\%] & F1 & Acc [\%] & Acc [\%] & Acc [\%] \\
    \midrule
    1 & 37.96 & 89.95 & 69.42 & 70.94 & 64.54 \\
    3 & 28.95 & 92.28 & 71.64 & 72.83 & 68.43 \\
    5 & \bf 26.31 & \bf 92.83 & \bf 72.79 & \bf 73.67 & \bf 69.92 \\
    \bottomrule
  \end{tabular}
  \end{footnotesize}
  \end{center}
}

\newcommand{\insertTabNLI}{
  \begin{center}
  \begin{footnotesize}
  \addtolength{\tabcolsep}{-3pt}
  \begin{tabular}{c*{5}{c}}
    \toprule
    NLI & Tatoeba & BUCC & MLDoc & XNLI-en & XNLI-xx \\
    obj.& Err [\%] & F1 & Acc [\%] & Acc [\%] & Acc [\%] \\
    \midrule
            - & \bf 26.31 & 92.83 &     72.79 & 73.67 & \bf 69.92 \\
    $\times$1 &     26.89 & 93.01 & \bf 74.51 & 73.71 &    69.10 \\
    $\times$2 & 28.52 & \bf 93.06 & 71.90 & 74.65 & 67.75 \\
    $\times$3 & 27.83 & 92.98 & 73.11 & \bf 75.23 & 61.86 \\
    \bottomrule
  \end{tabular}
  \end{footnotesize}
  \end{center}
}

\newcommand{\insertTabNLangs}{
  \begin{center}
  \begin{footnotesize}
  \addtolength{\tabcolsep}{-4pt}
  \begin{tabular}{cc*{5}{c}}
    \toprule
    \multirow{2}{*}{\#langs}
      & WMT & BUCC & MLDoc & XNLI-en & XNLI-xx \\
      & Err [\%] & F1 & Acc [\%] & Acc [\%] & Acc [\%] \\
    \midrule
    All (93)  & \bf 0.54 &     92.83 &     72.79 & \bf{73.67} & \bf 69.92 \\
    Eval (18) &     0.59 & \bf 92.91 & \bf 75.63 &      72.99 &     68.84 \\ %
    \bottomrule
  \end{tabular}
  \end{footnotesize}
  \end{center}
}